\documentclass[runningheads]{llncs}

\usepackage{eccv}

\usepackage{eccvabbrv}

\usepackage{booktabs}
\usepackage{amsmath}
\usepackage{multirow}
\usepackage{pifont}

\usepackage[accsupp]{axessibility}  % Improves PDF readability for those with disabilities.

 \usepackage[breaklinks,colorlinks,citecolor=eccvblue]{hyperref}
\usepackage{orcidlink}

\newcommand{\boldparagraph}[1]{\noindent{\bf #1} }

\newcommand{\cmark}{\text{\ding{51}}}
\newcommand{\xmark}{\text{\ding{55}}}

\usepackage[dvipsnames]{xcolor}
\usepackage{comment}

\newcommand{\PAR}[1]{\vskip4pt \noindent{\bf #1~}}

\usepackage[acronym]{glossaries}
\makeglossaries{}
\newacronym{tab}{Tab.}{Table}
\newacronym{pq}{PQ}{Product Quantization}

\newcommand{\figref}[1]{Fig.~\ref{#1}}
\newcommand{\secref}[1]{Section~\ref{#1}}
\newcommand{\eqnref}[1]{Eq.~\eqref{#1}}
\newcommand{\tabref}[1]{Table~\ref{#1}}

\begin{document}
\emergencystretch 3em

\title{Medical Image Segmentation with SAM-generated Annotations} 

\titlerunning{Medical Image Segmentation with SAM-generated Annotations}

% Second Author\inst{2,3}\orcidlink{1111-2222-3333-4444} \and
% Third Author\inst{3}\orcidlink{2222--3333-4444-5555}}

\author{Iira H{\"a}kkinen\inst{1,2}\orcidlink{0009-0009-2225-6959} \and
Iaroslav Melekhov\inst{2}\orcidlink{0000-0003-3819-5280} \and
Erik Englesson\inst{3}\orcidlink{0000-0003-4535-2520} \and \\
Hossein Azizpour\inst{3}\orcidlink{0000-0001-5211-6388} \and Juho Kannala\inst{2,4}\orcidlink{0000-0001-5088-4041}\\
\medskip
$^1$University of Helsinki ~ $^2$Aalto University \\$^3$KTH Royal Institute of Technology  $^4$University of Oulu}

%\author{\equalcontrib\inst{1} \and
%\equalcontrib\inst{2} \and \inst{1} \and \\ \inst{2} \and }

\authorrunning{I. H{\"a}kkinen et al.}

%\institute{Anonymous Organization\\\email{lncs@springer.com}}
\institute{}

\maketitle
%\renewcommand{\thefootnote}{\fnsymbol{footnote}}
%\footnotetext[2]{\dag~Equal contribution}
%\def\thefootnote{*}\footnotetext{denotes equal contribution. \\ Correspondence to: \email{zakaria.nits@gmail.com}, \email{iaroslav.melekhov@aalto.fi}}

\begin{abstract}
  The field of medical image segmentation is hindered by the scarcity of large, publicly available annotated datasets. Not all datasets are made public for privacy reasons, and creating annotations for a large dataset is time-consuming and expensive, as it requires specialized expertise to accurately identify regions of interest (ROIs) within the images. To address these challenges, we evaluate the performance of the Segment Anything Model (SAM) as an annotation tool for medical data by using it to produce so-called ``pseudo labels'' on the Medical Segmentation Decathlon (MSD) computed tomography (CT) tasks. The pseudo labels are then used in place of ground truth labels to train a UNet model in a weakly-supervised manner.
 %To alleviate these challenges, we experiment with the Segment Anything Model (SAM) to see if it has the potential to reduce the annotation workload required to generate labelled medical segmentation datasets. We autogenerate prompts to simulate an expert annotator, producing so-called ``pseudo labels'' on SAM. The pseudo labels are then used to train a UNet model, whose performance is compared against a fully supervised UNet model.
 We experiment with different prompt types on SAM and find that the bounding box prompt is a simple yet effective method for generating pseudo labels. This method allows us to develop a weakly-supervised model that performs comparably to a fully supervised model.
\end{abstract}

\section{Introduction}
%- Medical image segmentation, and its unique challenges in terms of annotation.
The field of medical image segmentation (MIS) faces challenges due to the scarcity of large, publicly available annotated datasets. The process of annotating segmentation masks is both time-consuming and expensive, typically requiring the expertise of medical professionals to accurately identify regions of interest (ROIs) within  images~\cite{mazurowski_segment_2023,wang_medical_2022,zhang_challenges_2024}. This challenge is one reason for the limited amount of publicly available medical datasets \cite{wang_medical_2022,zhang_challenges_2024}. 
%Not all datasets are made public for privacy reasons

%- What foundational models are, and how we could use them as annotation tools.
Foundational models have been proposed as an interesting way to deal with this challenge. Foundation models are large general models trained on massive, diverse datasets, and have been proven to be effective across multiple domains ~\cite{huix_are_2023}. As the field of MIS still relies on task-specific supervised models ~\cite{ma_segment_2024,zhang_challenges_2024}, foundation models could be used as data annotation tools to accelerate the labelling process of ground truth (GT) data ~\cite{huang_segment_2024,mattjie_zero-shot_2023,mazurowski_segment_2023}. Depending on the quality of the zero-shot predictions of the computer vision foundation model, the segmentations could either be accepted as is, or fine-tuned manually by experts.
%
%Using a CV foundation model for zero-shot label generation could accelerate this process, thereby enabling the annotation of a larger volume of data within a fixed time frame. Depending on the output of the foundation model, the segmentations could either be accepted as is, or fine-tuned manually by experts.
%While the general zero-shot performance of current computer vision foundation models on medical images can remain limited, techniques such as transfer learning and model fine-tuning demonstrate promising results ~\cite{huix_are_2023,zhang_challenges_2024}.

%- SAM.
In particular, one recent foundation model for semantic segmentation is the Segment Anything Model (SAM) developed by Meta AI ~\cite{kirillov_segment_2023}. SAM is a promptable encoder-decoder model trained on 11 million images along with over 1 billion masks, on mostly natural images. Along with the input image, a prompt can be given to SAM to indicate the region of interest (ROI). The prompt can be \eg, a point, or a bounding box around the ROI. SAM also provides a mode called everything mode, where the model is prompted to segment everything from the image with an even grid of point prompts over the input image. With the image and the prompt, the SAM model predicts one or several segmentation masks.

%- What others have done with SAM as an annotation tool and the research gap.
Using SAM as a data annotation tool has been studied in the literature~\cite{huang_segment_2024,ma_segment_2024,mattjie_zero-shot_2023,mazurowski_segment_2023}. These studies have compared the predictions with different prompting methods for SAM against ground truth labels. However, to the best of our knowledge, the final performance of models trained on SAM-generated labels has not been studied. This is crucial, as the distribution of the labels produced by SAM could be different from real labels, which can impact the model's training process and final performance. As the SAM-generated labels inevitably incorporate new noise into the training data, it is essential to evaluate the final model performance to assess the complete impact of the noisy labels.

%- What we do (contributions)
Without taking doctors away from patients, we propose to get as close to a real setting as possible by i) simulating using SAM as an annotation tool for medical images, and ii) evaluating the generalization ability of neural networks trained on these SAM-generated labels. We experiment with different prompt methods and train two sets of UNet models: fully supervised UNet trained with ground truth labels, and UNet trained with the SAM-generated pseudo labels using the simple and effective box prompt. The experiments are conducted on six abdominal medical segmentation tasks. The results show that the SAM model has great potential as a data annotation tool for medical images, and encourages further experimentation.

%In our approach, we use SAM with box prompt to produce so-called ''pseudo labels'', \ie labels to be used in place of ground truth labels when training a supervised medical image segmentation model. We compare two sets of UNet models: fully supervised UNet trained with ground truth labels, and UNet trained with the SAM-generated pseudo labels. The experiments are conducted on six abdominal medical segmentation tasks. The results show that the Segment Anything Model has great potential as a data annotation tool for medical images, and encourages further experimentation.
\section{Related Work}

Several studies have evaluated the zero-shot performance of SAM in the medical domain~\cite{gutierrez_no_2024,huang_segment_2024,ji_sam_2023,ma_segment_2024,mattjie_zero-shot_2023,mazurowski_segment_2023,roy_sammd_2023,zhang_segment_2024}. Researchers agree that SAM's performance in medical image segmentation tasks varies significantly depending on the specific task~\cite{huang_segment_2024,ji_sam_2023,zhang_segment_2024}. SAM performs well when ROI is large and distinct from the background but struggles with smaller, more complex structures. These challenges become more pronounced with ambiguous prompts, as the chosen prompting strategy affects SAM's final segmentation performance~\cite{gutierrez_no_2024,huang_segment_2024,roy_sammd_2023,zhang_segment_2024}. Additionally, SAM's performance is influenced by image qualities such as imaging modality, image dimensions, and contrast~\cite{huang_segment_2024,mazurowski_segment_2023,zhang_segment_2024}.

Huang~\etal~\cite{huang_segment_2024}, Ma~\etal~\cite{ma_segment_2024}, Wu~\etal~\cite{wu_medical_2023}, Zhang~\etal~\cite{zhang_segment_2024}, and Zhang and Liu~\cite{zhang_customized_2023} all report that for SAM to achieve state-of-the-art performance, the model needs to be fine-tuned for medical data. One significant factor contributing to the performance gap between SAM and other state-of-the-art medical segmentation approaches is the nature of the images in the SA-1B dataset~\cite{kirillov_segment_2023} used to train SAM. The SA-1B dataset consists mostly of natural images, which typically have distinct edges between objects and backgrounds, and are quite different from medical images~\cite{zhang_segment_2024}. Moreover, since SAM rarely meets state-of-the-art standards in its default mode (the Everything Mode)~\cite{huang_segment_2024,ji_sam_2023,zhang_segment_2024}, it is clear that educated human input in the form of prompts is necessary for SAM to perform well in the medical domain.

Recent methods have suggested using SAM as a data annotation tool for medical data~\cite{huang_segment_2024,mattjie_zero-shot_2023,mazurowski_segment_2023}. Both~\cite{mattjie_zero-shot_2023,mazurowski_segment_2023} propose that SAM could accelerate the process of creating ground truth labels if data annotators used SAM as a tool. Additionally, Huang et al.~\cite{huang_segment_2024} report that SAM could help human annotators by reducing the time needed for a single annotation while maintaining high quality. The authors conducted a study with three doctors who had to annotate 620 masks consisting of 55 different objects covering 9 medical imaging modalities.  The doctors performed annotations in two ways: a) from scratch and b) by adjusting labels predicted by SAM. The study found that using SAM improved annotation speed by approximately 25\% compared to fully manual annotation. Furthermore, the SAM approach achieved higher annotation quality, as measured by the Human Correction Efforts (HCE) index. The HCE index estimates the human effort required to fix predictions to reach a required accuracy level. While these studies~\cite{huang_segment_2024,mattjie_zero-shot_2023,mazurowski_segment_2023} have thoroughly evaluated different prompting strategies for SAM to produce medical labels, to the best of our knowledge, the final performance of models trained on SAM-generated labels has not been experimented on. We believe that assessing such models is crucial, especially in the medical domain, as it would provide a deep understanding of their generalization performance and it could also facilitate further research on self-supervised methods for semantic segmentation.    %While these studies  have thoroughly evaluated different prompting strategies for SAM to produce medical labels, to our knowledge, the final performance of models trained on SAM-generated labels has not been experimented on. % why important
\begin{figure}[t!]
\begin{center}
\centerline{\includegraphics[width=\columnwidth]{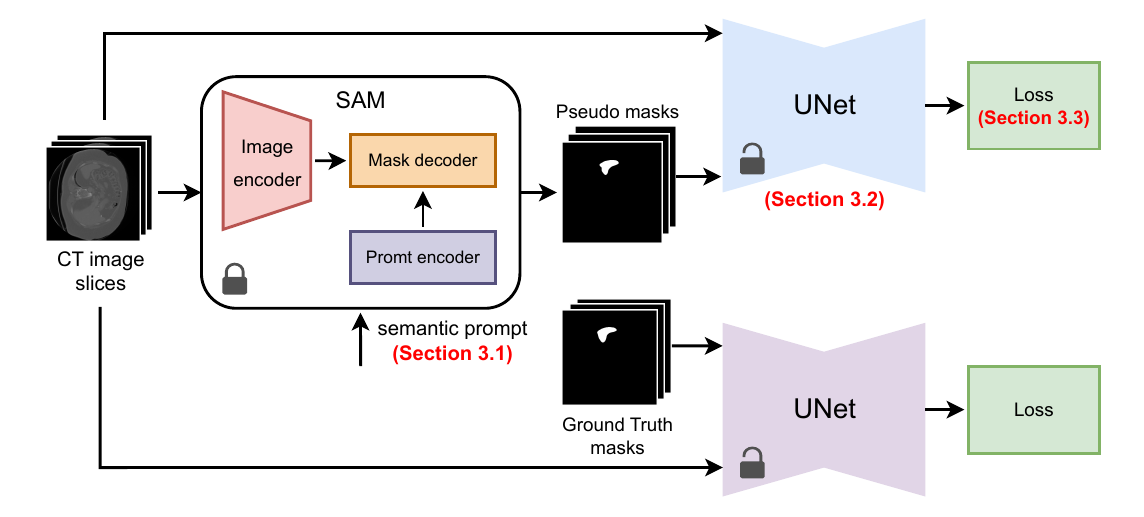}}
\caption{\textbf{Pipeline}. A set of 2D CT scans is propagated through the pre-trained SAM model~\cite{kirillov_segment_2023} to obtain the corresponding pseudo segmentation masks (\cf~\secref{ssec:label-gen}). Two independent UNet models are then trained (\cf~\secref{ssec:unet-training} and~\secref{ssec:optimization}) using ground truth and pseudo labels to perform semantic segmentation.}
\label{train-test-pipeline}
\end{center}
\end{figure}

\section{Method}
Our objective is to assess the performance of the SAM model~\cite{kirillov_segment_2023} as an annotation tool for medical images and perform semantic segmentation in a weakly-supervised manner using masks produced by SAM. The full procedure is illustrated in~\figref{train-test-pipeline}. Next, we go into details of the pseudo labels generation process (\secref{ssec:label-gen}), the weakly-supervised semantic segmentation (\secref{ssec:unet-training}), and the objective function (\secref{ssec:optimization}).
%To this end, we propose the following pipeline illustrated in~\figref{fig:prompt-examples}.
%
\subsection{Simulated Semantic Annotations Using SAM}\label{ssec:label-gen}
In this work, the SAM model is employed to generate semantic annotations by simulating expert input. Specifically, we consider the following prompting methods based on available ground-truth masks: a) $\text{Point}_{\text{CM}}$: a single positive point that represents the approximate center (center of mass, CM) of the region of interest (ROI); b) $\text{Point}_{\text{interior}}$: a single positive point that is furthest from the ground-truth edges of ROI; c) Box: a bounding box prompt is placed to \textit{tightly} enclose ROI; d) $\text{Box}_{\text{noise}}$: a bounding box prompt is placed to \textit{loosely} enclose ROI; e) $\text{Box}_{\text{+PP/NP}}$: a Box prompt followed by an interactive positive or negative point using $\text{Point}_{\text{interior}}$. This last prompting method simulates using SAM as an iterative annotation tool by using the resulting segmentation mask from the Box prompt to choose where to place and what type of point (positive/negative) to use. Suppose the largest component of false positives (in SAM prediction but not in GT) is larger than the largest component of false negatives (in GT but not in SAM prediction). In that case, a positive point is used, otherwise a negative, and the point is placed in the largest component using $\text{Point}_{\text{interior}}$.

We evaluate SAM with the aforementioned input prompts on the training split of the target dataset and report results in~\secref{sec:ablation-studies}~(\cf~\tabref{tab:sam-prompt-dice-scores}). Although the $\text{Box}_{\text{+PP/NP}}$ prompt leads to superior semantic segmentation performance, we follow~\cite{ma_segment_2024,mazurowski_segment_2023} and adopt the simple and effective Box prompt to generate pseudo semantic masks.

\subsection{Weakly-supervised Semantic Segmentation}\label{ssec:unet-training}

In our experiments, we first create pseudo labels on the training images using SAM with box prompt. Then, for each experimented segmentation task, two separate neural networks are trained on the training split of the dataset: a) a fully supervised network, and b) a network that uses the pseudo labels in place of ground truth labels during training. The training pipelines of the networks are exactly identical, except for the ground truth mask data used. We evaluate the networks on the testing split of the target dataset, and report results in~\secref{lbl:results}

\boldparagraph{Models.} We use the UNet architecture~\cite{ronneberger_u-net_2015} for our neural network in all our experiments due to its simplicity and widespread use in the field of medical segmentation \cite{hesamian_deep_2019,wang_medical_2022,zhang_challenges_2024}. 
% implementation details
The number of feature channels in our approach are 64, 128, 256, and 512, following the original UNet structure~\cite{ronneberger_u-net_2015}. Each convolution operation has a kernel size of 3, a stride of 1, and padding of 1. We incorporate batch normalization~\cite{ioffe_batch_2015} after each convolution to normalize data before the next step. The max-pooling layers and transposed convolutions are used for feature map downsampling and upsampling. The number of output layers of the network is $1$ to perform binary segmentation. The pseudo semantic masks are generated using the pre-trained SAM model~\cite{kirillov_segment_2023} with ViT-H backbone. We also experiment with MedSAM~\cite{ma_segment_2024}, a foundation model fine-tuned on the medical image domain, and report results in~\secref{sec:experiments}.

\subsection{Optimization}\label{ssec:optimization}
One of the criteria to optimize the network is to address class imbalance, a common issue in medical image segmentation where the region of interest may occupy a small portion of the image compared to the background. Specifically, we utilize the Dice loss which directly maximizes the overlap between the predicted segmentation and the ground truth and can be defined as follows:

\begin{equation}
\mathcal{L}_{\text{dice}} = 1 - \frac{2 \sum_i \hat{p}_i p_i}{\sum_i \hat{p}_i^2 + \sum_i p_i^2}
\end{equation}

\noindent where $\hat{p}_i$ and $p_i$ represent the predicted and the ground truth values, respectively. To further improve the overall pixel-level accuracy, we adapt the Cross-Entropy loss which measures the pixel-wise classification error and penalizes the model heavily for misclassifying individual pixels:

\begin{equation}
\mathcal{L}_{\text{ce}} = - \sum_i p_i \log(\hat{p}_i)
\end{equation}

The final loss is a weighted sum of the Dice and Cross-Entropy loss defined as:

\begin{equation}
\mathcal{L}_{\text{total}} = \mathcal{L}_{\text{dice}} + \alpha\mathcal{L}_{\text{ce}}
\end{equation}\label{eq:total_loss}

\noindent Combining these two losses~\cite{loss_odyssey,ma_segment_2024} provides a balance between the pixel-level accuracy and the region-level overlap. The Cross-Entropy loss helps in the initial stages of training by providing strong gradients, while the Dice loss becomes more important later, refining the segmentation boundaries and efficiently handling class imbalance.

\boldparagraph{Implementation details.} Before any evaluation or training, the pixel values of the 2D CT image slices are scaled to the range $[0, 1]$. For UNet training, additional data transformations are applied: from each training image, four random crops of size $1\times256\times256$ pixels are generated. These random crops are produced with a $2/3$ probability that the center pixel is classified as foreground in the ground truth data, and a $1/3$ probability that it is classified as background. Images smaller than $1\times256\times256$ pixels are padded to this size. The resulting images are then rotated by 90 degrees with a probability of $0.5$. All UNet models are trained for 50 epochs. After each epoch, the model is validated with the validation dataset and saved only if the average DSC metric on the validation set improves. We use a batch size of 2 and a learning rate of $5\cdot10^{-4}$ for all trained models. The AdamW optimizer~\cite{loshchilov_decoupled_2019} with the weight decay of $10^{-5}$ is employed for model optimization. During training, the cross-entropy term in~\eqnref{eq:total_loss} is scaled by $\alpha = 1$.
\begin{figure}[t!]
\begin{center}
\centerline{\includegraphics[width=\columnwidth]{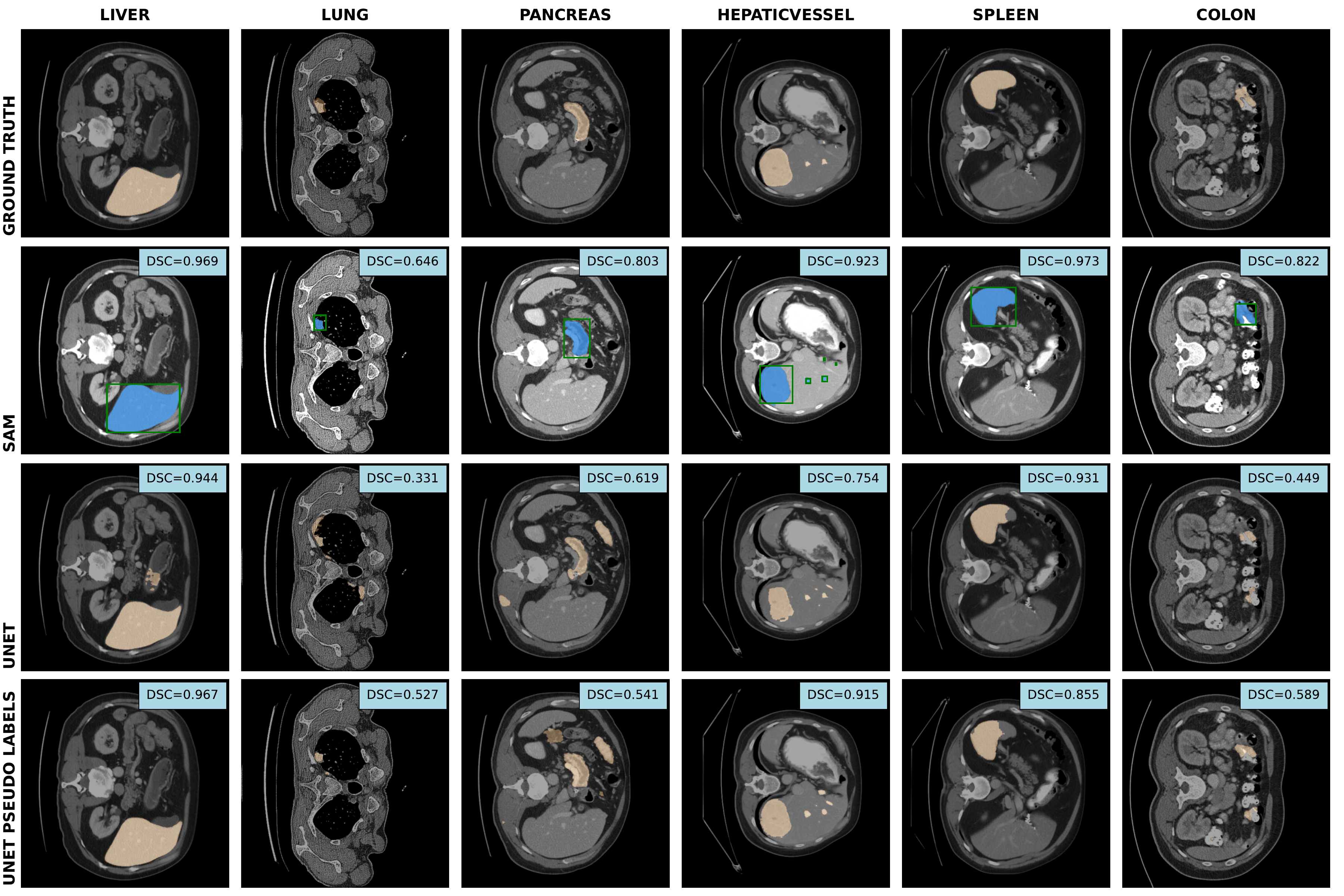}}
\caption{\textbf{Qualitative semantic segmentation results}. Each column displays one example case of each of the 6 segmentation tasks (from left to right: Liver, Lung, Pancreas, Hepatic Vessel, Spleen, Colon), and the rows from top to bottom display the ground truth label, SAM's predictions (with bounding box prompt), UNet's predictions, and predictions obtained by the UNet with pseudo labels.}
\label{example-segmentations}
\end{center}
\end{figure}

\section{Experiments}\label{sec:experiments}
We evaluate the proposed prompting methods and assess the performance of the SAM model on medical semantic segmentation and compare against state-of-the-art semantic segmentation models. We also provide an ablation of different prompt techniques of the proposed pipeline.

\boldparagraph{Dataset.} We use the Medical Segmentation Decathlon (MSD)~\cite{antonelli_medical_2022} dataset covering 10 segmentation tasks from the imaging modalities of computed tomography (CT), magnetic resonance imaging (MRI), and multiparametric magnetic resonance imaging (mp-MRI). In this work, we focus on the semantic segmentation of CT scans. The dataset consists of 6 different organs (tasks): Liver, Lung (tumor), Pancreas, Hepatic Vessels, Spleen, and Colon (tumor). Each organ is represented by corresponding training/test splits of 2D slices, as follows: Liver: 15429/3734; Lung: 1225/432; Pancreas: 6884/1908; Hepatic Vessels: 11053/1993; Spleen: 870/181; Colon: 1045/240.

\boldparagraph{Metric.} All models have been evaluated with the dice coefficient on the test split of each task, \ie $\text{DSC}=2TP/(2TP+FP+FN)$, where $TP$, $FP$, and $FN$ refer to the number of true positive, false positive, and false negative cases respectively.
%$\text{DSC} = 2 |A \cap B|/(|A| + |B|)$, where $A$ refers to the ground truth segmentation and $B$ refers to the model's predicted segmentation.
The calculations are done pixel-wise by comparing the model prediction with the corresponding ground truth mask from the test set.

\begin{table}[t!]
\caption{\textbf{Semantic Segmentation: Foundation Models.} We report the semantic segmentation results (the DSC metric, \cf~\secref{sec:experiments}) on each task/organ of the MSD dataset obtained by two foundation models, MedSAM~\cite{mazurowski_segment_2023} and SAM~\cite{kirillov_segment_2023}. The bounding box prompt is used for both models.} \label{tab:sam-vs-medsam}
\footnotesize
\centering % used for centering table
%\resizebox{\textwidth}{!}{
\setlength{\tabcolsep}{8.5pt}
\renewcommand*{\arraystretch}{1.05}
\begin{tabular}{l c c c c c c} 
 \toprule
 & & & & Hepatic & & \\ 
  & Liver & Lung & Pancreas & Vessels & Spleen & Colon \\ 
  \midrule
  MedSAM~\cite{ma_segment_2024} & \textbf{0.927} & 0.654 & 0.592 & 0.581 & 0.926 & 0.644 \\
  SAM~\cite{kirillov_segment_2023} & 0.926  & \textbf{0.734} & \textbf{0.836} & \textbf{0.752} & \textbf{0.932} & \textbf{0.754} \\
  %\midrule
  %SAM (everything mode) & 0.687 & 0.409 & 0.378 & 0.234 & 0.803 & 0.306 \\
 %UNet (GT labels) & 0.925 & 0.543 & 0.721 & 0.562 & 0.852 & 0.553 \\
 %UNet (pseudo labels) & 0.903 & 0.508 & 0.681 & 0.542 & 0.834 & 0.522 \\ 

 \bottomrule
\end{tabular}
%}
\end{table}

\begin{table}[t!]
\caption{\textbf{Semantic Segmentation on the Medical Segmentation Decathlon.} The DSC score is reported for each organ/task of the MSD dataset. The SAM* model is SAM evaluated in the Everything mode. Along with our models (UNet-based), we also test the recent fully-supervised medical semantic segmentation approaches.} \label{tab:main-results-msd}
\footnotesize
\centering % used for centering table
\setlength{\tabcolsep}{2.5pt}
\renewcommand*{\arraystretch}{1.05}
\begin{tabular}{l c c c c c c c c} 
 %\hline
 %\multicolumn{7}{|c|}{\textbf{Averaged Test Dice Scores}} \\
 \toprule
  & Sup. & Unsup. &  &  &  & Hepatic & & \\ 
  & Labels & Labels & Liver & Lung & Pancreas & Vessels & Spleen & Colon \\ 
  \midrule
  Trans VW~\cite{haghighi2021transferable} & \cmark & \xmark & 0.952 & 0.745 & 0.814 & 0.658 & 0.974 & 0.515 \\
  C2FNAS~\cite{c2fnas} & \cmark & \xmark & 0.950  & 0.704 & 0.808 & 0.643 & 0.963 & 0.589 \\
  Models Gen.~\cite{models-genesis} & \cmark & \xmark & 0.957 & 0.745 & 0.814 & 0.658  & 0.974 & 0.515 \\
  nnUNet~\cite{nn-unet} & \cmark & \xmark & 0.958 & 0.740 & 0.816 & 0.665 & 0.974 & 0.583 \\
  DiNTS~\cite{dints} & \cmark & \xmark & 0.954 & 0.748 & 0.810 & 0.645 & 0.970  & 0.592  \\
  Swin UNETR~\cite{swin} & \cmark & \xmark & 0.954 & 0.766 & 0.819 & 0.657 & 0.970 & 0.595 \\
  Universal model~\cite{CLIP-driven-universal} & \cmark & \xmark & 0.954 & 0.800 & 0.828 & 0.672 & 0.973 & 0.631 \\
  \midrule
  SAM* & \xmark & \xmark & 0.687 & 0.409 & 0.378 & 0.234 & 0.803 & 0.306 \\
  %\rowcolor{gray!10}
 UNet & \cmark & \xmark & 0.925 & 0.543 & 0.721 & 0.562 & 0.852 & 0.553 \\
 %\rowcolor{gray!10}
 UNet & \xmark & \cmark & 0.903 & 0.508 & 0.681 & 0.542 & 0.834 & 0.522 \\

 \bottomrule
\end{tabular}
%}
\end{table}

\subsection{Results\label{lbl:results}}
\boldparagraph{SAM vs. MedSAM.} We first evaluate two foundation models, SAM~\cite{kirillov_segment_2023} and MedSAM~\cite{mazurowski_segment_2023} on the test splits of the MSD dataset in the zero-shot setting. The MedSAM~\cite{mazurowski_segment_2023} follows the network architecture of SAM but is fine-tuned on a very large and diverse medical dataset consisting of CT, magnetic resonance imaging (MRI), and optical coherence tomography (OCT) data. We use the bounding box prompt for both models and report results in~\tabref{tab:sam-vs-medsam}. Interestingly, the SAM model outperforms its more advanced domain-specific counterpart, except for the Liver. We assume that while using more diverse and specialized medical training data can improve performance for MedSAM, it can also limit the generalization ability compared to a more broadly trained model like SAM.

\boldparagraph{Supervised- vs. Weakly-supervised Methods.} \tabref{tab:main-results-msd} presents the averaged test dice scores for UNet and UNet with pseudo labels, along with the zero-shot SAM results and state-of-the-art semantic segmentation models. To compute the pseudo labels, the box prompt was chosen as the prompting method for SAM through ablation studies on different prompt types (\cf~\secref{sec:ablation-studies}). The box prompt provides consistent results with test dice scores ($\text{DSC} \geq 0.734$) across the six tasks of various sizes and complexity, while being a relatively simple prompting strategy. The experimental results indicate that UNet with pseudo labels performs comparably to the fully-supervised baseline, UNet, with the absolute difference in test Dice score ranging from $0.018$ (Spleen) to $0.04$ (Pancreas). Interestingly, the weakly-supervised UNet even performs on par with state-of-the-art models on several tasks/organs such as Liver, Colon, and Spleen. Notably, the UNet-based models demonstrate modest performance on some tasks. Specifically, the Lung, Spleen, and Colon have the smallest training dataset sizes (\cf~\secref{sec:experiments}, Datasets) that could partially explain UNet's weak Dice score results for the Lung and Colon tasks. However, UNet achieves good accuracy for the Spleen, despite having the smallest training dataset among the six tasks.  The Spleen is a large organ with a homogeneous form and clear boundaries, whereas the Lung and Colon tasks involve tumors that often have outlines difficult for even experts to agree on which is in line with previous study~\cite{antonelli_medical_2022}. Therefore, in addition to dataset sizes, the characteristics of the tasks seem to affect the final test Dice scores of UNet. In~\figref{example-segmentations}, we qualitatively examine the SAM-generated labels and the predictions of the UNet models on examples from the six tasks.

\begin{table}[t!]
\caption{\textbf{Ablation on different prompt types.} SAM's zero-shot dice score performance on the training set with different prompt types. CM refers to the centre of mass, and +PP/ NP refers to an additional positive or negative point.}
\label{tab:sam-prompt-dice-scores}
\footnotesize
\centering
\renewcommand*{\arraystretch}{1.05}
\setlength{\tabcolsep}{8pt}
\begin{tabular}{l c c c c c c} 
 \toprule
 &  &  &  & Hepatic &  &  \\  
Prompt & Liver & Lung & Pancreas & Vessel & Spleen & Colon \\ 
  \midrule
 $\text{Point}_\text{CM}$ & 0.818 & 0.500 & 0.580 & 0.130 & 0.863 & 0.480 \\
 $\text{Point}_\text{interior}$ & 0.830 & 0.500 & 0.593 & 0.130 & 0.865 & 0.525 \\
 $\text{Box}$ & 0.927 & 0.713 & 0.822 & 0.735 & 0.923 & 0.782 \\
 $\text{Box}_{\text{noise}}$ & 0.913 & 0.677 & 0.774 & 0.499 & 0.907 & 0.704 \\
 $\text{Box}_{\text{+PP/NP}}$ & \textbf{0.935} & \textbf{0.774} & \textbf{0.844} & \textbf{0.743} & \textbf{0.931} & \textbf{0.808} \\
 \bottomrule
\end{tabular}
\end{table}

\subsection{Ablation Studies\label{sec:ablation-studies}}

The need for prompts was validated by running SAM with Everything Mode on the test splits of the tasks (\cf~\tabref{tab:main-results-msd}). As the Everything Mode outputs multiple masks for one input image, the final mask was chosen as the best dice score match with the ground truth mask of the test image in question, which adds supervision to the approach. The averaged test dice scores on everything mode are reported in~\tabref{tab:main-results-msd}, and range from $0.234$ (Hepatic Vessels) to $0.803$ (Spleen). The SAM model with Everything Mode struggled to segment small structures and ROIs with unclear edges. While this approach could achieve promising dice scores for some of the large and clearly outlined organs, the good performance on such tasks was not consistent.

\begin{figure}[t!]
\begin{center}
\centerline{\includegraphics[width=\columnwidth]{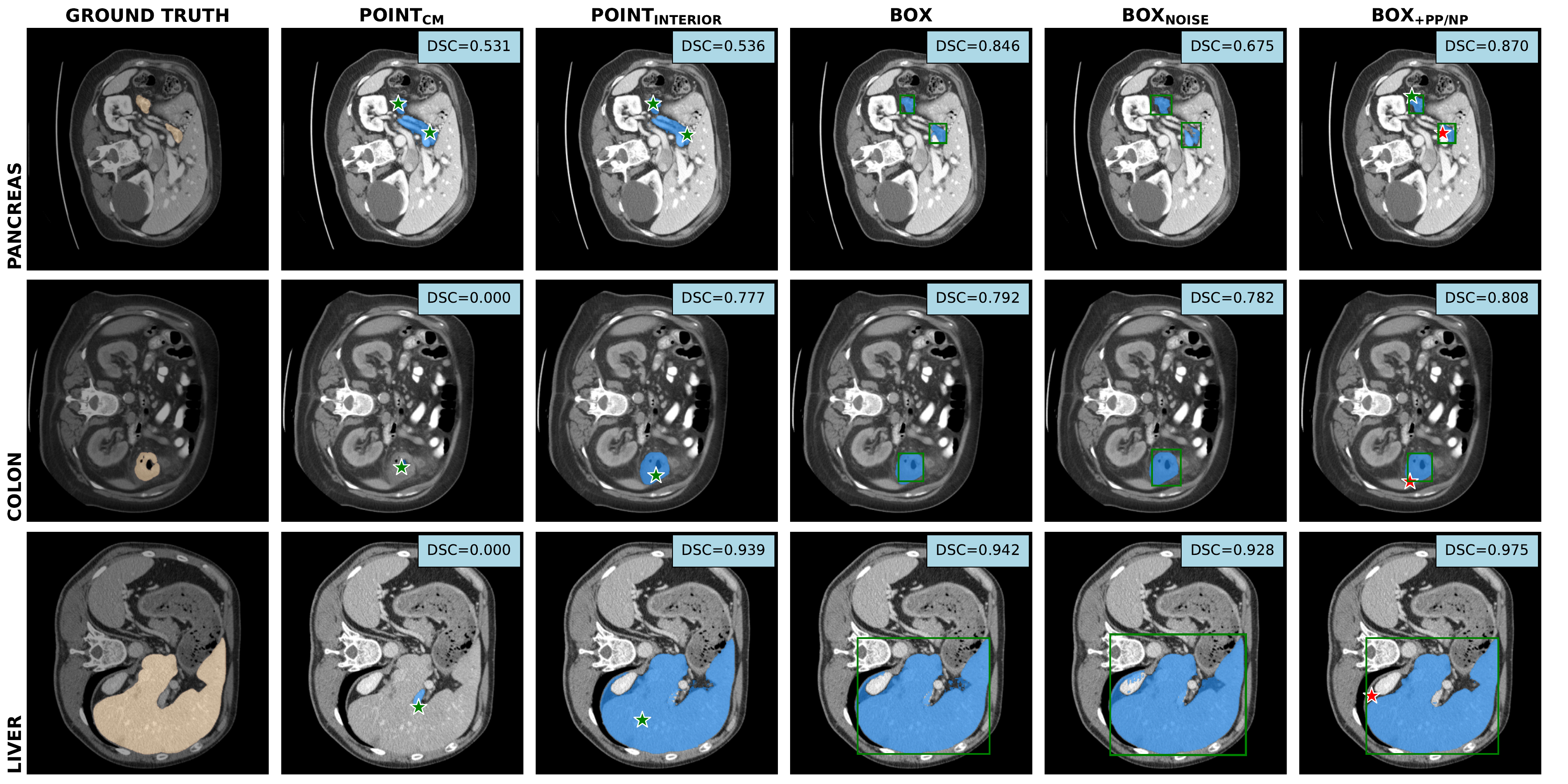}}
\caption{\textbf{Ablation on different prompt types}. Segmentation masks from different prompts. Green and red stars are positive and negative points, respectively.} 
\label{fig:prompt-examples}
\end{center}
\end{figure}

The results in~\tabref{tab:sam-prompt-dice-scores} show the importance of different prompts when using SAM as an annotation tool. We find that point prompts generally perform worse than box prompts. This is highlighted especially on the more ambiguous tasks, \eg, hepatic vessels, where the mask foreground consists of multiple small, irregular clusters. These experiments confirm observations done in previous literature ~\cite{huang_segment_2024,ji_sam_2023,zhang_segment_2024}, where it was suggested that SAM can achieve great performance on larger, homogeneous structures, such as liver and spleen, but struggles more on heterogeneous, complex structures. However, combining the box prompt with an additional positive or negative point achieves the highest test dice scores on all datasets. In \figref{fig:prompt-examples}, we present qualitative results for the different prompt methods on example CT scans. As the simple box prompt offers strong performance with less work for annotators compared to $\text{Box}_\text{+PP/NP}$, it was chosen as the prompting method for the pseudo label generation. 

\section{Conclusion}

To conclude, our experiments show positive results on the possibility of using SAM as a data annotation tool for medical images, and encourages further experiments on the subject. Most significantly, our findings indicate that a model trained on SAM-generated pseudo labels can achieve performance comparable to that of a fully supervised model. We note that the prompting strategy has a significant effect on pseudo label accuracy. While SAM's bounding box pseudo label accuracy can be further improved with additional iterative points, the experiments show that a simpler prompting method, \ie box prompt, is still able to generate consistently reliable pseudo labels across tasks of various sizes and complexities. This observation is noteworthy, as a simple and efficient prompting method allows for faster data annotation, while maintaining final model performance comparable to fully supervised training.

\PAR{Acknowledgments}
This work was partially supported by the Wallenberg AI, Autonomous Systems and Software Program (WASP) funded by the Knut and Alice Wallenberg Foundation. Some of the experiments were performed using the supercomputing resource Berzelius provided by the National Supercomputer Centre at Linköping University and the Knut and Alice Wallenberg foundation. JK also acknowledges funding from the Academy of Finland (grant No. 327911, 353138). We acknowledge CSC -- IT Center for Science, Finland, and the Aalto Science-IT project for computational resources.

\bibliographystyle{splncs04}
\bibliography{main}

\begin{thebibliography}{10}
\providecommand{\url}[1]{\texttt{#1}}
\providecommand{\urlprefix}{URL }
\providecommand{\doi}[1]{https://doi.org/#1}

\bibitem{antonelli_medical_2022}
Antonelli, M., Reinke, A., Bakas, S., Farahani, K., Kopp-Schneider, A., Landman, B.A., Litjens, G., Menze, B., Ronneberger, O., Summers, R.M., van Ginneken, B., Bilello, M., Bilic, P., Christ, P.F., Do, R.K.G., Gollub, M.J., Heckers, S.H., Huisman, H., Jarnagin, W.R., {McHugo}, M.K., Napel, S., Pernicka, J.S.G., Rhode, K., Tobon-Gomez, C., Vorontsov, E., Meakin, J.A., Ourselin, S., Wiesenfarth, M., Arbeláez, P., Bae, B., Chen, S., Daza, L., Feng, J., He, B., Isensee, F., Ji, Y., Jia, F., Kim, I., Maier-Hein, K., Merhof, D., Pai, A., Park, B., Perslev, M., Rezaiifar, R., Rippel, O., Sarasua, I., Shen, W., Son, J., Wachinger, C., Wang, L., Wang, Y., Xia, Y., Xu, D., Xu, Z., Zheng, Y., Simpson, A.L., Maier-Hein, L., Cardoso, M.J.: The medical segmentation decathlon. Nature Communications  \textbf{13}(1) (2022)

\bibitem{gutierrez_no_2024}
Guti{\'e}rrez, J.D., Rodriguez-Echeverria, R., Delgado, E., Rodrigo, M.{\'A}.S., S{\'a}nchez-Figueroa, F.: No more training: {SAM}’s zero-shot transfer capabilities for cost-efficient medical image segmentation. {IEEE} Access  \textbf{12},  24205--24216 (2024)

\bibitem{haghighi2021transferable}
Haghighi, F., Taher, M., Zhou, Z., Gotway, M.B., Liang, J.: Transferable visual words: Exploiting the semantics of anatomical patterns for self-supervised learning. In: IEEE Transactions on Medical Imaging (2021)

\bibitem{dints}
He, Y., Yang, D., Roth, H., Zhao, C., Xu, D.: {DiNTS}: Differentiable neural network topology search for 3d medical image segmentation. In: Proceedings of the IEEE/CVF Conference on Computer Vision and Pattern Recognition (CVPR). pp. 5841--5850 (2021)

\bibitem{hesamian_deep_2019}
Hesamian, M.H., Jia, W., He, X., Kennedy, P.: Deep learning techniques for medical image segmentation: Achievements and challenges. Journal of Digital Imaging  \textbf{32}(4),  582--596 (2019)

\bibitem{huang_segment_2024}
Huang, Y., Yang, X., Liu, L., Zhou, H., Chang, A., Zhou, X., Chen, R., Yu, J., Chen, J., Chen, C., Liu, S., Chi, H., Hu, X., Yue, K., Li, L., Grau, V., Fan, D.P., Dong, F., Ni, D.: Segment anything model for medical images? Medical Image Analysis  \textbf{92},  103061 (2024)

\bibitem{huix_are_2023}
Huix, J.P., Ganeshan, A.R., Haslum, J.F., Söderberg, M., Matsoukas, C., Smith, K.: Are natural domain foundation models useful for medical image classification? arXiv:2310.19522  (2023)

\bibitem{ioffe_batch_2015}
Ioffe, S., Szegedy, C.: Batch normalization: Accelerating deep network training by reducing internal covariate shift. In: International Conference on Machine Learning. pp. 448--456. {PMLR} (2015)

\bibitem{nn-unet}
Isensee, F., Jaeger, P.F., Kohl, S.A., Petersen, J., Maier-Hein, K.H.: {nnU-net}: A self-configuring method for deep learning-based biomedical. Nature Methods  \textbf{18},  203--211 (2021)

\bibitem{ji_sam_2023}
Ji, G.P., Fan, D.P., Xu, P., Zhou, B., Cheng, M.M., Van~Gool, L.: {SAM} struggles in concealed scenes — empirical study on “segment anything”. Science China Information Sciences  \textbf{66}(12),  226101 (2023)

\bibitem{kirillov_segment_2023}
Kirillov, A., Mintun, E., Ravi, N., Mao, H., Rolland, C., Gustafson, L., Xiao, T., Whitehead, S., Berg, A.C., Lo, W.Y., Dollár, P., Girshick, R.: Segment anything. arXiv:2304.02643  (2023)

\bibitem{CLIP-driven-universal}
Liu, J., Zhang, Y., Chen, J.N., Xiao, J., Lu, Y., A~Landman, B., Yuan, Y., Yuille, A., Tang, Y., Zhou, Z.: Clip-driven universal model for organ segmentation and tumor detection. In: Proceedings of the IEEE/CVF International Conference on Computer Vision (ICCV). pp. 21152--21164 (2023)

\bibitem{loshchilov_decoupled_2019}
Loshchilov, I., Hutter, F.: Decoupled weight decay regularization. arXiv:1711.05101  (2019)

\bibitem{loss_odyssey}
Ma, J., Chen, J., Ng, M., Huang, R., Li, Y., Li, C., Yang, X., Martel, A.L.: Loss odyssey in medical image segmentation. Medical Image Analysis  \textbf{71} (2021)

\bibitem{ma_segment_2024}
Ma, J., He, Y., Li, F., Han, L., You, C., Wang, B.: Segment anything in medical images. Nature Communications  \textbf{15}(1) (2024)

\bibitem{mattjie_zero-shot_2023}
Mattjie, C., De~Moura, L.V., Ravazio, R., Kupssinskü, L., Parraga, O., Delucis, M.M., Barros, R.C.: Zero-shot performance of the segment anything model ({SAM}) in 2d medical imaging: A comprehensive evaluation and practical guidelines. In: International Conference on Bioinformatics and Bioengineering (BIBE). pp. 108--112 (2023)

\bibitem{mazurowski_segment_2023}
Mazurowski, M.A., Dong, H., Gu, H., Yang, J., Konz, N., Zhang, Y.: Segment anything model for medical image analysis: An experimental study. Medical Image Analysis  \textbf{89} (2023)

\bibitem{ronneberger_u-net_2015}
Ronneberger, O., Fischer, P., Brox, T.: {U-Net}: Convolutional networks for biomedical image segmentation. In: Proceedings of the Medical Image Computing and Computer Assisted Intervention – MICCAI. pp. 234--241. Springer International Publishing (2015)

\bibitem{roy_sammd_2023}
Roy, S., Wald, T., Koehler, G., Rokuss, M.R., Disch, N., Holzschuh, J., Zimmerer, D., Maier-Hein, K.H.: {SAM}.{MD}: Zero-shot medical image segmentation capabilities of the segment anything model. arXiv:2304.05396  (2023)

\bibitem{swin}
Tang, Y., Yang, D., Li, W., Roth, H.R., Landman, B., Xu, D., Nath, V., Hatamizadeh, A.: Self-supervised pre-training of swin transformers for 3d medical image analysis. In: Proceedings of the IEEE/CVF Conference on Computer Vision and Pattern Recognition (CVPR). pp. 20730--20740 (2022)

\bibitem{wang_medical_2022}
Wang, R., Lei, T., Cui, R., Zhang, B., Meng, H., Nandi, A.K.: Medical image segmentation using deep learning: A survey. {IET} Image Processing  \textbf{16}(5),  1243--1267 (2022)

\bibitem{wu_medical_2023}
Wu, J., Ji, W., Liu, Y., Fu, H., Xu, M., Xu, Y., Jin, Y.: Medical {SAM} adapter: Adapting segment anything model for medical image segmentation. arXiv:2304.12620  (2023)

\bibitem{c2fnas}
Yu, Q., Yang, D., Roth, H., Bai, Y., Zhang, Y., Yuille, A.L., Xu, D.: {C2FNAS}: Coarse-to-fine neural architecture search for 3d medical image segmentation. In: IEEE/CVF Conference on Computer Vision and Pattern Recognition (CVPR). pp. 4126--4135 (2020)

\bibitem{zhang_customized_2023}
Zhang, K., Liu, D.: Customized segment anything model for medical image segmentation. arXiv:2304.13785  (2023)

\bibitem{zhang_challenges_2024}
Zhang, S., Metaxas, D.: On the challenges and perspectives of foundation models for medical image analysis. Medical Image Analysis  \textbf{91} (2024)

\bibitem{zhang_segment_2024}
Zhang, Y., Shen, Z., Jiao, R.: Segment anything model for medical image segmentation: Current applications and future directions. Computers in Biology and Medicine  \textbf{171} (2024)

\bibitem{models-genesis}
Zhou, Z., Sodha, V., Pang, J., Gotway, M.B., Liang, J.: Models genesis. Medical Image Analysis  \textbf{67} (2021)

\end{thebibliography}
\end{document}